\documentclass{article}

\usepackage{iclr2027_conference,times}

\usepackage{amsmath,amsfonts,bm}

\def\eqref#1{equation~\ref{#1}}
\def\Eqref#1{Equation~\ref{#1}}

\def\1{\bm{1}}

\DeclareMathAlphabet{\mathsfit}{\encodingdefault}{\sfdefault}{m}{sl}
\SetMathAlphabet{\mathsfit}{bold}{\encodingdefault}{\sfdefault}{bx}{n}

\usepackage{microtype}
\usepackage{graphicx}
\usepackage{subcaption}
\usepackage{booktabs}
\usepackage{hyperref}
\usepackage{url}
\usepackage{amsmath}
\usepackage{amssymb}
\usepackage{mathtools}
\usepackage{amsthm}
\usepackage{xcolor}
\usepackage{comment}
\usepackage{multirow}
\usepackage{tabularx}
\usepackage{placeins}
\usepackage{algorithm}
\usepackage{algorithmic}
\usepackage[capitalize,noabbrev]{cleveref}

\newcommand{\RComment}[1]{\hfill$\triangleright$~#1}
\newcommand{\ignore}[1]{}

\theoremstyle{plain}
\newtheorem{theorem}{Theorem}[section]
\newtheorem{proposition}[theorem]{Proposition}

\theoremstyle{definition}

\theoremstyle{remark}

\title{Variability Aware Recursive Neural Network (VARNN): A Residual-Memory Model for Capturing Temporal Deviation in Sequence Regression Modeling}

\author{
\textbf{Haroon Gharwi\textsuperscript{1*} \quad Yue Dai\textsuperscript{2*} \quad Kai Shu\textsuperscript{3+}}\\[3pt]
\textsuperscript{*}Department of Computer Science, Illinois Institute of Technology, Chicago, IL, USA \\
\textsuperscript{+}Department of Computer Science, Emory University, Atlanta, GA, USA\\[3pt]
{\small
\textsuperscript{1}\texttt{hgharwi@hawk.illinoistech.edu} \quad
\textsuperscript{2}\texttt{ydai21@illinoistech.edu} \quad
\textsuperscript{3}\texttt{kai.shu@emory.edu}
}}

\iclrfinalcopy  

\makeatletter
\renewcommand{\@thanks}{}
\renewcommand{\@oddhead}{}
\renewcommand{\@evenhead}{}
\makeatother

\usepackage{fancyhdr}
\fancypagestyle{varnnstyle}{
  \fancyhf{} 
  \rfoot{\small Preprint. © 2026} 
  \cfoot{\thepage}               
  \renewcommand{\headrulewidth}{0pt}
  \renewcommand{\footrulewidth}{0pt}
}
\begin{document}

\maketitle

\
\begin{abstract}
Real-world time-series regression often involves non-stationarity,
heteroscedasticity, and regime changes, under which recent prediction
errors may contain structured information about local temporal mismatch
between model predictions and observations. Learning how to represent
and reuse these errors can therefore provide useful information for
subsequent prediction.
We introduce the Variability-Aware Recursive Neural Network
(\textsc{VARNN}), a residual-aware architecture for supervised
time-series regression that learns an explicit residual-memory state
from recent prediction errors and uses it to condition subsequent
predictions. Specifically, \textsc{VARNN} maps scalar prediction
innovations into a learned nonlinear, vector-valued residual
representation over a short context.
Across nine datasets spanning energy, healthcare, and environmental
domains, VARNN achieves lower test MSE than the compared static,
lag-based, and sequence-model baselines. Targeted ablations further show
that learned projected residual memory improves predictive accuracy over
direct scalar residual feedback, supporting the benefit of a learned
nonlinear representation of prediction deviations.
\end{abstract}

\section{Introduction}
Regression is one of the fundamental tasks in machine learning, where the goal is to learn a mapping from input features to a continuous-valued output. Classical regression models, such as linear regression, assume a static (stationary) relationship between predictors and target, whereas flexible learners \ignore{while more advanced methods}(e.g., random forests and neural networks) capture nonlinear dependencies \cite{hyndman2021fpp3}. In many real-world applications, however, observations are not independent and identically distributed (i.i.d.) but instead arrive as time series, where observations are temporally ordered and often correlated. \cite{brockwell2002introTS}.

This motivates dynamic time series regression, where the objective is to predict the output at each timestep using the current covariates and a short history of recent observations (one-step-ahead forecasting, per-timestep supervised prediction) \cite{hyndman2021fpp3,brockwell2002introTS}, where historical observations can reveal temporal information that is not fully explained by the current covariates alone.
In the machine learning literature, A common approach is to represent temporal information explicitly through lagged observations. Closely related formulations appear as Autoregressive with Exogenous Input (ARX) and Nonlinear ARX (NARX) models in system identification and control \cite{ljung1999systemid,lin1996narx}.
Such prediction problems arise across energy analytics, biomedical monitoring,
and environmental modeling, where local temporal dynamics interact with
exogenous drivers \cite{candanedo2017appliances,pimentel2016rr,zhang2017beijing}.
However, standard lagged-regression approaches primarily model dependence
on past inputs and outputs and do not explicitly represent how prediction
errors themselves evolve over time.

Prediction errors can be particularly informative when the underlying
process exhibits non-stationarity, heteroscedasticity, transient
disturbances, or regime changes. Under such conditions, recent residuals
may contain structured information about local mismatch between the
learned predictor and the evolving data-generating process rather than
behaving purely as independent noise.\ignore{Classical ARMAX models explicitly
address this issue by augmenting autoregressive dynamics with past
innovations. } Specifically, ARMAX represents error dynamics through a pre-specified linear moving-average combination of lagged scalar-valued
innovations \cite{pankratz2012forecasting}. While this provides an explicit
error-correction mechanism and can capture linear autocorrelation in the
innovation process, the representation remains structurally restricted:
the residuals are scalar-valued and their temporal influence is governed
by a fixed linear form. However, real-world prediction deviations can arise from multiple, potentially nonlinear sources of variability,
including drift, heteroscedasticity, transient disturbances, and regime changes \cite{hyndman2021fpp3,brockwell2002introTS}. A scalar innovation summarizes the net prediction mismatch at a given step, but does not provide a flexible representation
through which a predictor can respond differently to these heterogeneous error patterns. Thus, a fixed linear representation of lagged scalar innovations, as used in ARMAX, may be restrictive when prediction deviations exhibit heterogeneous and potentially nonlinear patterns of variability.

Another dynamic approach is to learn temporal representations directly
from sequences. Recurrent neural networks (RNNs), together with variants
such as LSTMs and GRUs, maintain a latent state that is updated recursively,
allowing the model to summarize historical information and capture
nonlinear temporal dependencies over time
\cite{lipton2015rnnreview,hochreiter1997lstm,cho2014gru}. Compared with
fixed lag-based models, this learned latent state provides substantially
greater representational flexibility. Nevertheless, standard recurrent
models typically represent variability only implicitly: prediction
residuals primarily influence learning through the optimization objective
rather than being maintained as a dedicated stateful source of information.
Consequently, the hidden state must simultaneously summarize input
dynamics, historical context, and any error-related structure.

To address this gap, we introduce the
\emph{Variability-Aware Recursive Neural Network}
(\textsc{VARNN}), a residual-aware neural regression framework
\footnote{The term ``VARNN'' in this work denotes a
\textit{Variability-Aware Recursive Neural Network}. It is unrelated to
the classical ``Vector Autoregressive (VAR)'' statistical model used in
econometrics.}.
\textsc{VARNN} explicitly promotes the one-step innovation
$e_\tau = y_\tau-\hat{y}_\tau$ from a scalar prediction error to a
learned nonlinear, vector-valued \emph{residual-memory} as main recurrent state
$\mathbf{h}_\tau$. This state is then fused with the covariates to
condition subsequent predictions. This enables the model to adapt
to non-stationary conditions, offering greater robustness across diverse time series domains such as
energy, biomedical, and environmental systems.

Our main contributions are as follows:  
\begin{enumerate}
    \item We introduce VARNN, a compact residual-memory architecture that promotes the one-step innovation from a scalar prediction error to an explicit nonlinear, latent vector-valued
    residual-memory state used to condition subsequent predictions.

    \item We systematically compare VARNN against three families of regression baselines: static, dynamic lag and sequence, including ARX, ARMAX, MLP, RNN, Dlinear and PatchTST.  
    \item We demonstrate through experiments on across nine datasets spanning energy, healthcare, and environmental domains. that VARNN achieves superior robustness. Targeted ablations further examine residual-memory update mechanisms.
\end{enumerate}

\section{Related Work}

\textbf{Static regression.} Classical regression methods, including linear regression, ridge, lasso, kernel models, and tree ensembles, are sample-efficient and interpretable but treat each $(\mathbf{x}_t,y_t)$ independently. They lack mechanisms to encode temporal dependence or adapt to time-varying uncertainty, leading to degraded accuracy under autocorrelation and non-stationarity \citep{hyndman2021fpp3, degooijer2006twentyfive}. 

\textbf{Dynamic lag regression.} ARX and their nonlinear counterparts NARX incorporate lagged outputs and inputs, offering practical interpretability. However, they typically assume white, homoscedastic innovations and require careful lag-order selection, which limits robustness under distribution shift \citep{pankratz2012forecasting, diversi2010identification, clark2020modern}. Neural extensions improve flexibility while preserving interpretability \citep{dong2025endtoend}. The ARMAX adds moving average terms to capture serially correlated innovations, improving fit when shocks persist but still relying on fixed parametric assumptions and pre-specified error horizons.

\textbf{Sequence models.} RNNs, LSTMs, and GRUs learn temporal dependencies by maintaining latent states, but they absorb variability into hidden dynamics without distinguishing systematic structure from stochastic disturbances, leading to instability under non-stationarity \citep{hochreiter1997lstm, cho2014gru, lipton2015rnnreview, lim2021deeplearning}. Beyond RNNs, Temporal Convolutional Networks and Transformer-style forecasters extend receptive fields or apply attention mechanisms, with mixed robustness under distribution shift \citep{zeng2023dlinear, bai2018tcn, lim2019tft, zhou2021informer, oreshkin2019nbeats}.

\textbf{Non-stationarity-aware methods.}
Existing approaches address temporal variability through
different mechanisms. RevIN normalizes and restores
instance statistics to mitigate distribution shift
\citep{kim2022revin}, while neural volatility hybrids
incorporate conditional-variance modeling
\citep{han2024garch}. Other architectures explicitly
address non-stationarity in time-series prediction
\citep{baidya2024nonstationarity,liu2025nonstationarity}.
VARNN focuses on a different mechanism: learning a
nonlinear residual-memory representation from observed
prediction errors to condition subsequent predictions.

\begin{figure*}[!t]
    \centering
    \includegraphics[width=0.8\linewidth]{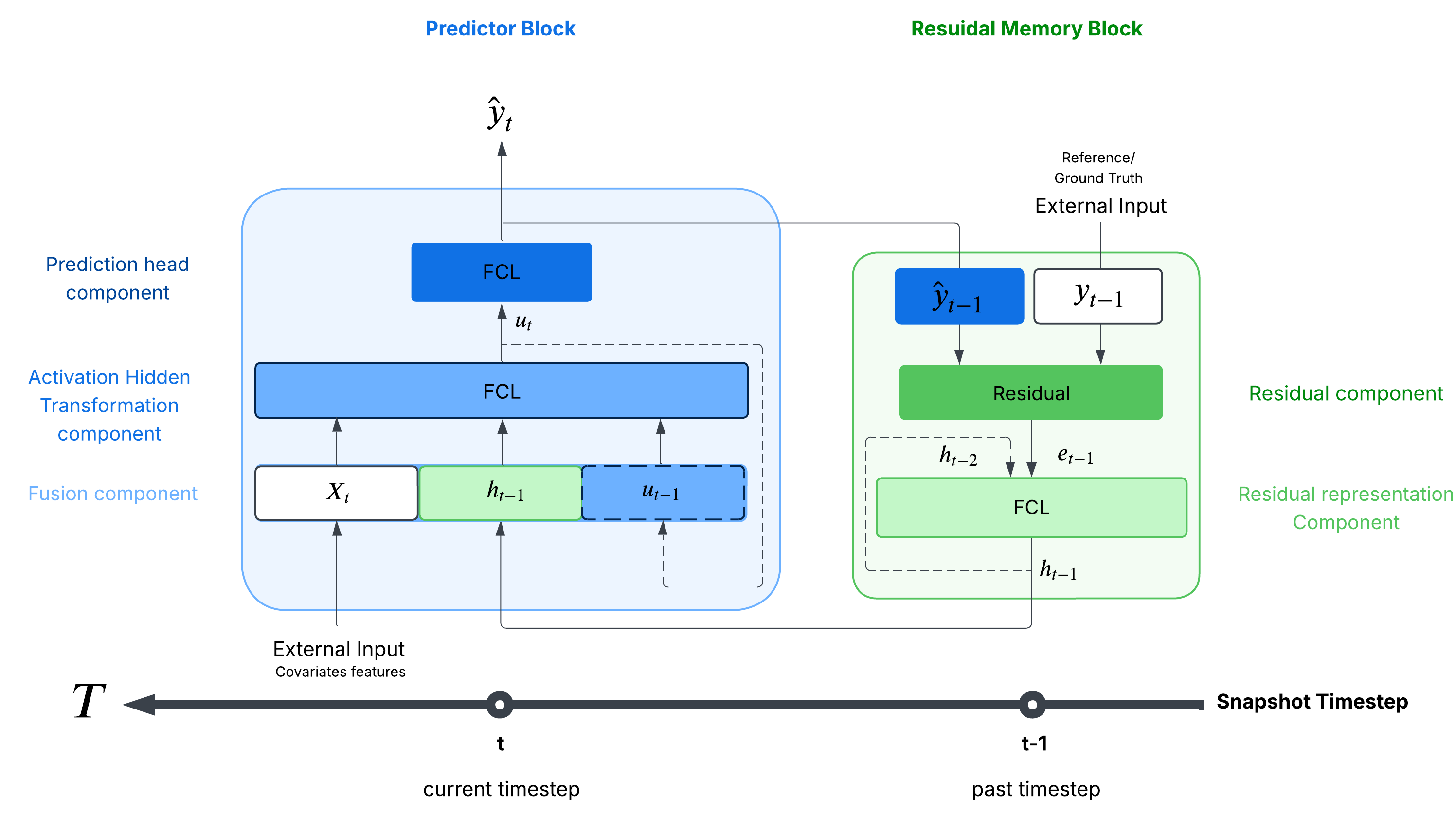}
    \caption{VARNN architecture: the Residual Memory Block (green) converts the innovation $e_{\tau}$ into a residual-memory state $\mathbf{h}_{\tau}$; the Predictor Block (blue) fuses $\mathbf{x}_\tau$ with $\mathbf{h}_{\tau-1}$ and, when enabled, $\mathbf{u}_{\tau-1}$, to produce $\hat y_\tau$.}
    \label{fig:VARNN}
\end{figure*}

\section{Methodology}
\label{sec:methodology}
\subsection{Problem Definition}
We consider supervised time–series regression with multivariate covariates and a univariate response:

$\{(\mathbf{x}_\tau, y_\tau)\}_{\tau=1}^{T}$, where $\mathbf{x}_\tau \in \mathbb{R}^d,\; y_\tau \in \mathbb{R}.$
For each prediction time $t$, we construct a sliding window of length $w$:
\begin{equation}
\mathcal{W}_t := \{(\mathbf{x}_{t-w+1}, y_{t-w+1}), \ldots, (\mathbf{x}_{t-1}, y_{t-1}), \mathbf{x}_t\},
\label{eq:window}
\end{equation}
where $y_t$ is unknown at prediction time. The goal is to predict the target $\hat{y}_t$ from the past $w{-}1$ labeled steps and the current covariates such that:
\begin{equation}
\hat{y}_t \;=\; f_\theta\!\big(\mathbf{x}_{t-w+1:t},\; y_{t-w+1:t-1}\big).
\label{eq:map}
\end{equation}
The learning objective is to estimate the model parameters \(\boldsymbol{\theta}\) by minimizing the mean squared error (MSE) over the \(N\) training sampling windows: $\mathcal{L}(\boldsymbol{\theta}) \;=\; \frac{1}{N}\sum_{t=1}^{N} \bigl(y_t - \hat{y}_t\bigr)^2$.

\subsection{Model Architecture}
\label{sec:varnn}
Variability-Aware Recursive Neural Network (VARNN) is a dynamic sequence regressor that explicitly encodes recent prediction errors into the next-step computation via a dedicated residual-memory pathway. Unlike classical RNNs, which update hidden states solely from inputs and prior activations, VARNN promotes the one-step innovation $e_\tau$ to a \emph{residual-memory} state that calibrates the next prediction, making \emph{prediction innovations} \ignore{error} the primary\ignore{first-class } recurrent signal. As shown in \autoref{fig:VARNN}, the architecture comprises two coupled blocks unrolled over the $w{-}1$ labeled context steps and used once more for the horizon prediction at time $t$:

\paragraph*{\ignore{Inference/prediction}Predictor Block}
The Predictor Block’s purpose is to \emph{map fused inputs to the latent representation $\mathbf{u}_\tau$ and produce the one-step prediction $\hat y_\tau$}. Given $\mathbf{x}_\tau \in \mathbb{R}^d$, residual memory state $\mathbf{h}_{\tau-1} \in \mathbb{R}^m$, we form the Fusion component as following: 
\begin{equation}
\mathbf{z}_\tau \;=\; [\,\mathbf{x}_\tau;\,\mathbf{h}_{\tau-1}\,], 
\qquad \mathbf{z}_\tau \in \mathbb{R}^{d+m}.
\label{eq:fusion}
\end{equation}

We compute a latent representation\ignore{activation} by mapping\ignore{transforms} the fused vector $\mathbf{z}_\tau$ to a $k$-dimensional activation through a nonlinear layer:
\begin{equation}
\mathbf{u}_\tau \;=\; \sigma(\mathbf{W}_z \mathbf{z}_\tau + \mathbf{b}_z), \qquad 
\mathbf{u}_{\tau} \in \mathbb{R}^k
\label{eq:activation}
\end{equation}
where $\sigma(\cdot)$ is a pointwise nonlinearity (ReLU by default). This transformation is then projected to a linear scalar to estimate  $\hat{y}_\tau$ as one–step prediction:
\begin{equation}
\hat{y}_\tau \;=\; \mathbf{W}_o \mathbf{u}_\tau + \mathbf{b}_o, \qquad 
\hat{y}_{\tau} \in \mathbb{R}
\label{eq:prediction}
\end{equation}

Concretely, at each supervised step $\tau$ ($\tau\!\le\!t{-}1$), inputs $\mathbf{x}_\tau$ are fused with the previous residual memory $\mathbf{h}_{\tau-1}$, mapped to the latent representation, and read out to a prediction. It recursively implements the forward computation for $\hat y_\tau$ while refreshing the residual memory block.

\paragraph*{Residual Memory Block:}
The Residual Memory block is an error-aware memory that encodes how prediction deviates from the current observation as a signal of variability and stores it as state that conditions the next step. The one-step prediction error is a reusable state that conditions the next step. Given the current supervised step prediction $\hat{y}_\tau$ and ground truth ${y}_\tau$ ,we form the innovation \ignore{residual error/ innovation/ prediction error}:
\begin{equation}
e_\tau \;=\; y_\tau - \hat{y}_\tau \;\in\; \mathbb{R}.
\label{eq:innovation}
\end{equation}

This \ignore{residual error} innovation scalar $e_\tau $ is embedded into an $m$-dimensional residual-memory representation via a learnable projection and a pointwise nonlinearity activation function denoted as $\rho(\cdot)$. We update the memory state by:
\begin{equation}
\mathbf{h}_{\tau} = \rho\!\left(\mathbf{W}_\epsilon e_{\tau} + \mathbf{b}_\epsilon\right).  
\qquad
\mathbf{h}_\tau \in \mathbb{R}^{m}.
\label{eq:memory}
\end{equation}

By updating the state with the innovation, we treat the residual as a \emph{primary recurrent input}. This \emph{explicit error pathway} exposed variability and misspecification to the next-step predictor through the fusion term in \eqref{eq:fusion}. 
The \emph{Predictor Block} performs the transformation of the input fusion and prediction; the \emph{Residual-Memory Block} handles error assimilation. Together, they form a lightweight recurrence framework that emphasizes variability tracking. 




\paragraph{Current-time prediction}
At current time step $t$, we predict $\hat{y_t}$ using the current covariates $x_t$ and last available memory state without performing a residual update:
\begin{equation}
\hat{y}_t \;=\; \mathbf{W}_o^\top\,
\sigma\!\big(\mathbf{W}_z\,[\,\mathbf{x}_t;\,\mathbf{h}_{t-1}\,] + \mathbf{b}_z\big) \;+\; b_o .
\label{eq:current_time_pred}
\end{equation}

Once $y_t$ is observed\ignore{becomes available} (offline or streaming), the residual memory can be refreshed\ignore{it can be used to update the memory with $e_t=y_t-\hat y_t$ for the subsequent window} via \eqref{eq:memory} for subsequent windows.

\subsection{Training Objective and Optimization}
We minimize the mean–squared error (MSE) on the current step prediction:
\begin{equation}
\mathcal{L}(\theta)
=\frac{1}{N}\sum_{\text{windows}} \big(y_t-\hat{y}_t\big)^2,\quad
\theta=\{\mathbf{W}_z,\mathbf{b}_z,\mathbf{W}_o,\mathbf{b}_o,\mathbf{W}_\epsilon,\mathbf{b}_\epsilon\}.
\label{eq:loss_mse}
\end{equation}

\subsection{Model Variants}
\label{sec:variants}
While the core VARNN architecture defines a residual-memory pathway that encodes innovations into a state $\mathbf{h}_\tau$, different update dynamics could affect how variability is represented and reused. Therefore, we study several update rules and augmentations that trade off responsiveness and stability. We instantiate four variants along two orthogonal design axes: (i) whether the residual memory update is \emph{instantaneous} or \emph{accumulative}, and (ii) whether the model also carries forward an \emph{activation memory} $\mathbf{u}_{\tau-1}$ in addition to the residual state. This yields the four following configurations:

\textbf{VARNN-RM} (\emph{Residual Memory only}): 
    This is the base model that uses the instantaneous update in \eqref{eq:memory}, where the innovation $e_\tau$ is projected directly into the memory state. This lightweight design reacts only to the most recent error. This provides a sharp correction based solely on the most recent deviation. \ignore{suitable for signals with abrupt but transient variability, suitable when variability is abrupt but short-lived. provide the minimal parameters}
    
\textbf{VARNN-RM+AM} (\emph{Residual + Activation Memory}): 
    In addition to residual memory, the previous latent activation $\mathbf{u}_{\tau-1}$ is carried forward into the fusion vector in \eqref{eq:fusion}. The fused component of predictor input becomes $\mathbf{z}_\tau = [\,\mathbf{x}_\tau;\,\mathbf{h}_{\tau-1};\,\mathbf{u}_{\tau-1}\,],$ allowing the predictor to integrate a compact nonlinear summary of recent fused inputs alongside the residual state. \ignore{improving representation of short-term dependencies.} 
    
\textbf{VARNN-ARM} (\emph{Accumulative Residual Memory}): 
    which accumulates recent innovations into a smoother state. This variant behaves like a learned nonlinear autoregressive filter on prediction errors. In residual representation component \eqref{eq:memory}, we replace the instantaneous update with an accumulative rule by:
    \begin{equation}
    \mathbf{h}_\tau = \rho\!\left(W_h \mathbf{h}_{\tau-1} + W_\epsilon e_\tau + b_\epsilon\right), \qquad
    \mathbf{h}_\tau \in \mathbb{R}^{m}.
    \label{eq:memory-ARM}
    \end{equation}
    which introduces persistence and enables the residual-memory state to integrate multiple past innovations. This design is motivated by settings with drift or systematic bias, where cumulative error provides a corrective signal.
    
\textbf{VARNN-ARM+AM} (\emph{Accumulative Residual + Activation Memory}): 
    Combines accumulative residual memory with activation carry-over, yielding the richest variant. It retains nonlinear summaries of fused inputs while integrating error dynamics over time, aimed at regimes with both strong short-term variability and slow distributional shifts.

These four variants highlight different ways to leverage error information: RM emphasizes responsiveness, ARM emphasizes persistence, AM augments representational capacity, and their combinations balance the two. We empirically compare all variants in Section~\ref{sec:results} to assess which design choices best enhance robustness under non-stationary conditions.

\subsection{Computational Complexity}
\label{sec:complexity}
Let $d$ be the covariate dimension, $k$ the predictor width, and $m$ the residual–memory width (default $m{=}d$).
Relative to a covariate-only feed-forward predictor with fusion weight $W_z\!\in\!\mathbb{R}^{k\times d}$, the base model \textbf{VARNN-RM}
concatenates $\mathbf{h}_{\tau-1}\!\in\!\mathbb{R}^m$, expanding $W_z$ to $\mathbb{R}^{k\times(d+m)}$ and adding $km$ parameters
and $km$ multiplies per step. The residual projection $W_r\!\in\!\mathbb{R}^{m\times 1}$ adds $m$ parameters and $m$ multiplies.
Ignoring biases, the incremental overhead is therefore $O(km)$ parameters and multiplies. By contrast, a standard RNN update
$\mathbf{h}_\tau=\rho(W_x\mathbf{x}_\tau + W_h\mathbf{h}_{\tau-1})$ costs $O(dk + k^2)$ per step. Since $m\!\ll\!k$ in practice,
the residual pathway adds negligible overhead relative to an RNN cell. \textit{Note: the linear readout head is shared across models and is omitted from these counts.}

\subsection{Algorithmic View}
Algorithm~\ref{alg:varnn1} summarizes the teacher-forced unroll over the $(w{-}1)$ context steps and the current-time prediction for the base model \textbf{VARNN-RM}. For all other variants, see \autoref{app:alg-view}
    
\begin{algorithm}[!t]
\caption{VARNN: Windowed prediction at time $t$}
\label{alg:varnn1}
\tiny 
\begin{algorithmic}
\REQUIRE Window $\mathcal{W}_t=\{(\mathbf{x}_{t-w+1},y_{t-w+1}),\ldots,(\mathbf{x}_{t-1},y_{t-1}),\mathbf{x}_t\}$; parameters $\theta=\{\mathbf{W}_z,\mathbf{b}_z,\mathbf{W}_o,\mathbf{b}_o,\mathbf{W}_r,\mathbf{b}_r\}$
\ENSURE Current prediction $\hat{y}_t$

\STATE \textbf{Init residual memory:} $\mathbf{h}_{t-w}\gets \mathbf{0}$
\FOR[teacher-forced context (supervised) steps]{$\tau=t-w+1$ \textbf{to} $t-1$}
\STATE $\mathbf{z}_\tau \gets [\,\mathbf{x}_\tau;\,\mathbf{h}_{\tau-1}\,]$
    \STATE \textbf{Inference:} $\mathbf{u}_\tau \gets \theta(\mathbf{W}_z\mathbf{z}_\tau+\mathbf{b}_z)$; \quad $\hat{y}_\tau \gets \mathbf{W}_o\mathbf{u}_\tau+\mathbf{b}_o$
    \STATE \textbf{Innovation:} $e_\tau \gets y_\tau-\hat{y}_\tau$
    \STATE \textbf{Residual update:} $\mathbf{h}_\tau \gets \rho(\mathbf{W}_r e_\tau+\mathbf{b}_r)$
\ENDFOR
\STATE \RComment{current-time prediction (no residual update at $t$)}
\STATE  $\mathbf{z}_t \gets [\,\mathbf{x}_t;\,\mathbf{h}_{t-1}\,]$
\STATE  $\hat{y}_t \gets \mathbf{W}_o\, \sigma(\mathbf{W}_z\mathbf{z}_t+\mathbf{b}_z)+\mathbf{b}_o$
\STATE \textbf{return} {$\hat{y}_t$ }
\end{algorithmic}

\end{algorithm}

\subsection{VARNN vs Standard Recurrent Models (RNN/LSTM/ GRU)}
Conventional recurrent models update a hidden state from $(\mathbf{x}_\tau,\mathbf{h}_{\tau-1})$ and learn to accommodate variability implicitly through gradients. VARNN makes the \emph{innovation process} a first-class citizen by dedicating a pathway (RM/ARM) that injects prediction error back into the state, explicitly targeting drift, heteroscedasticity, and short-horizon miscalibration.

\section{Experiment Settings}
\label{sec:exp}

We evaluate our proposed model \textbf{VARNN} on multivariate sequence-to-one regression (also called one step ahead forecasting) across three domains (energy, healthcare, environment). The goal is to test whether explicit residual-memory yields robustness under non-stationarity relative to srong static, dynamic (lagged), and sequence baselines, under identical preprocessing, splits, and training protocols.


\paragraph{Datasets.}
We use nine popular datasets grouped by domain summarized in the Table~\ref{tab:data-summary}. All tasks are scalar regression per time step as following: 

\textbf{Energy.}
    (i) \textbf{Appliances Energy Prediction} (\emph{Appliances}) records household power consumption with indoor/outdoor weather and occupancy covariates; the target is appliance energy use (Wh) \citep{candanedo2017appliances}.
    (ii) \textbf{ETTh1} and (iii) \textbf{ETTh2} are the electricity transformer temperature benchmarks with hourly electricity load and meteorological features to predict the oil temperature from exogenous covariates \citep{zhou2021informer}.
\textbf{Healthcare.}
    We use three scalar targets derived from the BIDMC PPG/ECG recordings \citep{pimentel2016rr}: (iv) \textbf{BIDMC HR} (heart rate), (v) \textbf{BIDMC RR} (respiratory rate), and (vi) \textbf{BIDMC SPO2} (oxygen saturation). In all cases, the covariates are three channels short-term waveform statistics and auxiliary signals  (PPG, RESP, ECG) at 125 Hz recorded from 52 patients. \ignore{and the response is a continuous physiological index.}
\textbf{Environmental.}
    (vii) \textbf{Beijing PM2.5} and (viii) \textbf{Beijing PM10} contain air-quality and meteorological variables over 12 monitoring stations; we predict PM$_{2.5}$ and PM$_{10}$ concentrations respectively \citep{zhang2017beijing}.
    (ix) \textbf{WEATHER} is a multi-variate meteorological time series where the target is the weather temperature single scalar \citep{zhou2021informer}. \ignore{details and feature lists are provided in Appendix~\ref{app:data-details}}

\ignore{ Dataset summary after prepossessing: sample size, missingness,
feature dimension $d$, and scalar prediction target. Although BIDMC and
Beijing originate from unified data sources, we evaluate each target
variable as a distinct supervised dataset, yielding nine datasets across
three domains.}

\begin{table}[t]
\centering
\caption{Dataset summary (post-cleaning): size, missingness, feature count $d$, and scalar target.}
\label{tab:data-summary}

\tiny
\setlength{\tabcolsep}{4pt}
\renewcommand{\arraystretch}{1}

\begin{tabular}{l l c c c l}
\toprule
\textbf{Domain} &
\textbf{Dataset} &
\textbf{Size} &
\textbf{Missing} &
\textbf{\#Feat. $(d)$} &
\textbf{Target} \\
\midrule

\multirow{3}{*}{Energy}
& Appliances  & 19{,}735  & No  & 27  & Appliance energy use (Wh) \\
& ETTh1       & 17{,}420  & No  & 6   & Oil temperature \\
& ETTh2       & 17{,}420  & No  & 6   & Oil temperature \\
\midrule

\multirow{3}{*}{Healthcare}
& BIDMC--HR   & 25{,}436  & No  & 375 & Heart rate (bpm) \\
& BIDMC--RR   & 25{,}436  & No  & 375 & Respiratory rate (breaths/min) \\
& BIDMC--SPO2 & 25{,}436  & No  & 375 & Oxygen saturation (SpO$_2$) \\
\midrule

\multirow{3}{*}{Environmental}
& Beijing--PM2.5 & 420{,}768 & Yes & 9 & PM$_{2.5}$ concentration \\
& Beijing--PM10  & 420{,}768 & Yes & 9 & PM$_{10}$ concentration \\
& WEATHER        & 35{,}064  & No  & 6 & Temperature ($^\circ$C) \\
\bottomrule

\end{tabular}
\end{table}

\paragraph{Preprocessing, windowing, and split.}
Continuous inputs and targets are min--max scaled using training statistics and applied unchanged to validation/test. We adopt sequence-to-one with window length $w{=}5$ and stride $1$: the first $w{-}1$ steps warm up residual memory; the model predicts the final step. Splits are chronological (80/20 train/test). Windows never cross split boundaries. For the PM2.5 and PM10 datasets, we filled missing values per station before scaling. Both are subject-wise splits (per station). BIDMC HR, RR, and SPO2 consist of three channels (PPG, RESP, ECG) at 125 Hz; we partition the signal into 1-second bins and concatenate the 125 samples per channel (no overlap), yielding 375 input vector features per time step. Each dataset is a subject-wise split (per patient).

\paragraph{VARNN Variants.}
We report two primary variants in the main table: \textbf{VARNN--RM}: instantaneous residual-memory update and \textbf{VARNN--RM+AM}: residual-memory + activation-memory fusion. The Two accumulative variants (ARM, ARM+AM) are included in the Appendix; they exhibit comparable accuracy with improved stability in highly variable environments.  We report the two primary variants (RM and RM+AM) in Table \ref{tab:main-results}. The two accumulative variants (ARM, ARM+AM) are included in the Appendix achieve comparable accuracy with improved stability, as detailed in Appendix \ref{app:pm25-variants}.

\paragraph{Baselines.}
We compare \textsc{VARNN} against thirteen baselines grouped into three broad regression families: static, dynamic lag, and dynamic sequence regression, all trained under identical preprocessing, scaling, and chronological splits. \textbf{(1) Static Regressors:} these models use only contemporaneous covariates $\mathbf{x}_t$ with no temporal history. They include Linear Regression (LR), Random Forest (RF), and a Multilayer Perceptron (MLP), each learning a direct mapping $f(\mathbf{x}_t)\rightarrow y_t$. \textbf{(2) Dynamic Lag Regressors:} These models incorporate explicit lag concatenation over the same horizon used by \textsc{VARNN} windowing ($w{=}5$). ARx-LR uses only lagged targets and current exogenous covariates $\mathbf{x}_t$, ARX-LR incorporates lagged covariates and lagged targets with current $\mathbf{x}_t$,  NARX-RF and NARX-MLP provide nonlinear autoregressive variants; and ARMAX-LR extends ARX with moving–average innovation terms. DLinear ~\citep{zeng2023dlinear} applies trend–seasonal decomposition with a linear head over lagged covariates and $\mathbf{x}_t$. \textbf{(3) Dynamic Sequence Regressors:} These models treat the windowed covariates $\mathbf{x}_{t-w+1:t}$ as a temporal sequence. We evaluate RNN, LSTM, and GRU, each predicting via a linear head on the final hidden state. We also evaluate PatchTST the SOTA transformer-based sequence baseline using a patching mechanism \cite{nie2023patchtst}. \ignore{Details of configuration settings are provided in Appendix \ref{app: hyperparameters}}

\paragraph{Hyperparameters, training, and libraries.}
\label{app: hyperparameters}
Neural models (VARNN, RNN/LSTM/GRU, and MLP) use hidden size $128$, ReLU activations, and are implemented in TensorFlow. SOTA models DLinear and PatchTST are implemented using the official PyTorch code from their original repositories; PatchTST is configured with embedding dimension $d_{\text{model}}{=}128$, feedforward dimension $128$, $4$ attention heads, one Transformer encoder layers, patch length $3$, stride $1$. ALL neural models are trained with a linear a one step regression head with Adam ($\mathrm{lr}=3{\times}10^{-3}$), batch size~$128$, for up to $50$ epochs with early stopping on validation MSE (patience $50$ with best–model restoration). Random forest models use $500$ trees with scikit–learn defaults. Linear regression baselines (LR, ARx--LR, ARX--LR) also use scikit–learn. The ARMAX is implemented via statsmodels library, with autoregressive and moving average orders chosen p and q=$4$ to match the same lag window and exogenous structure used by the other dynamic regression baselines. 

\paragraph{Evaluation Protocol.}
\label{sec:training-protocol}
We evaluated and reported the models based on the mean squared error (MSE) at the epoch with the best validation MSE. Data splits and parameter initializations are fixed under the same seed number (seed=2025). 

\section{Results}
\label{sec:results}
We evaluate models under the protocol described in Sec.~\ref{sec:exp} and report mean squared error (MSE; lower is better) on the training and test splits of all three datasets. Models are grouped into \emph{Static} baselines (no lag) and \emph{Dynamic} regressors (explicit lags or sequence). Table~\ref{tab:main-results} summarizes the results. \ignore{The two accumulative variants (ARM, ARM+AM) are included in the Appendix achieve comparable accuracy with improved stability, as detailed in Appendix \autoref{app:pm25-variants}.}

\begin{table*}[!t]
\centering
\caption{Train and test mean squared error (MSE; ↓) on all datasets. 
\emph{Static} models use only contemporaneous covariates $\mathbf{x}_t$; 
\emph{Dynamic} models additionally incorporate lagged outputs/inputs or recurrent states. 
Best test scores are in \textbf{bold}.}
\label{tab:main-results}
\setlength{\tabcolsep}{3pt}

\resizebox{\textwidth}{!}{%
\renewcommand{\arraystretch}{1.4}

\begin{tabular}{lcc|cc|cc|cc|cc|cc|cc|cc|cc}
\toprule

\multirow{3}{*}{\textbf{Model}} 
& \multicolumn{6}{c}{\textbf{ENERGY}}
& \multicolumn{6}{c}{\textbf{Healthcare}}
& \multicolumn{6}{c}{\textbf{Environmental}} \\
\cmidrule(lr){2-7}\cmidrule(lr){8-13}\cmidrule(lr){14-19}
& \multicolumn{2}{c}{\textbf{Appliances}} 
& \multicolumn{2}{c}{\textbf{ETTh1}}
& \multicolumn{2}{c}{\textbf{ETTh2}} 
& \multicolumn{2}{c}{\textbf{BIDMC HR}} 
& \multicolumn{2}{c}{\textbf{BIDMC RR}} 
& \multicolumn{2}{c}{\textbf{BIDMC SPO2}} 
& \multicolumn{2}{c}{\textbf{Beijing PM2.5}}
& \multicolumn{2}{c}{\textbf{Beijing PM10}}

& \multicolumn{2}{c}{\textbf{WEATHER}} 
\\
\cmidrule(r){2-3} \cmidrule(r){4-5} \cmidrule(r){6-7} \cmidrule(r){8-9} \cmidrule(r){10-11} \cmidrule(r){12-13}  \cmidrule(r){14-15}  \cmidrule(r){16-17} \cmidrule(r){18-19}
& \textbf{Train} & \textbf{Test} & \textbf{Train} & \textbf{Test} & \textbf{Train} & \textbf{Test} & \textbf{Train} & \textbf{Test} & \textbf{Train} & \textbf{Test} & \textbf{Train} & \textbf{Test} & \textbf{Train}   & \textbf{Test} & \textbf{Train}&\textbf{Test} & \textbf{Train} &\textbf{Test}\\
\midrule
\multicolumn{19}{l}{\emph{Static (no lags)}} \\
\hline
\quad \ignore{Linear Regression} LR        & 0.00799 & 0.00657       &0.02792 & 0.03335 & 0.02243 & 0.04112            & 0.01830 & 0.02256 &  0.00984 &  0.00822  &  0.03659   &  0.03103   & 0.00212 & 0.00171 &  0.00375   &  0.00242                           &  0.01516   &  0.01483   \\
\quad \ignore{Random Forest} RF            & 0.00057 & 0.04815       & 0.00234 & 0.02888 & 0.00174 & 0.03274            & 0.00031 & 0.01842 &  0.00032 &  0.00705  &   0.00057  &  0.03925   & 0.00012 & 0.00162 &   0.00025  &  0.00227                         &  0.00229   &  0.01235   \\
\quad \ignore{Multilayer Perceptron} MLP   & 0.00833 & 0.00778       & 0.00833 & 0.00778 & 0.02622 & 0.02049            & 0.00216 & 0.00168 & 0.01276  &  0.00933  & 0.04683    &  0.03452   & 0.00216 & 0.00168 &   0.00392  &   0.00224                          & 0.01413    &  1.21301   \\
\hline
\multicolumn{19}{l}{\emph{Dynamic (lags)}} \\
\hline
\quad Dlinear                       & 0.01317 & 0.01076              & 0.54476 & 0.30845 & 0.45580 & 0.43386           & 0.04563 & 0.10959 & 0.04307  &  0.08024  &   0.10516  &  0.12605   & 0.01346 & 0.01338 &   0.01477  &  0.01468                    &  0.39243   &  0.40353   \\
\quad ARx--LR                       & 0.00395 & 0.00958               & 0.00041 & 0.04819 & 0.00030 & 0.24168           & 0.00069 & 0.03362 & 0.00041  &  0.01834  &  0.00100  &  0.05868   & 0.00028 & 0.00274 &   0.00087  &  0.00290                     &  0.00042   &  0.34684   \\

\quad ARX--LR                       & 0.00610 & 0.00504               & 0.00067 & 0.00033 & 0.00076 & 0.00075           & 0.00023 & 0.00027 & 0.00030  &  0.00052  &  0.00028  &  0.00046   & 0.00051 & 0.00049 &   0.00144  &  0.00091                     &  0.00061   &  0.00069   \\
\quad NARX--RF                      & 0.00079 & 0.02750               & 0.00008 & 0.00036 & 0.00008 & 0.00085          & 0.00003 & 0.00035 & 0.00004  &  0.00053  &  0.00004   &  0.00068   & 0.00005 & 0.00049 &   0.00015  &   0.00093                   &  0.00006   &  0.00051   \\
\quad NARX--MLP                     & 0.00652 & 0.00527               & 0.00074 & 0.00036 & 0.00082 & 0.00089           & 0.00065 & 0.00079 & 0.01390  &  0.01381  &  0.00065   &  0.00098   & 0.00053 & 0.00052 &  0.00147   &  0.00094                   &  0.00079   & 0.00108    \\
\quad ARMAX-LR                       & 0.00411 & 0.01002              & 0.00038 & 0.02952 & 0.00027 & 0.14067         & 0.00086 & 0.03092 & 0.00050  &  0.01136  & 0.00132   & 0.04648   & 0.00027 & 0.06462 &  0.00084   &  0.00429                      &   0.00035  &  0.29126   \\
\hline
\multicolumn{19}{l}{\emph{Dynamic (Sequence)}} \\
\hline
\quad PatchTST                      & 0.00604 & 0.00689               &  0.01560 & 0.03251 & 0.03077 & 0.04006          & 0.01961 & 0.01900 & 0.00921  &  0.00878  &  0.05707   &  0.04502   & 0.00450 & 0.00492 &  0.00673   &   0.00621                  &  0.04191   &  0.03640   \\
\quad RNN                          & 0.00765 & 0.00697                & 0.01618 & 0.01779 & 0.01608 & 0.02246          & 0.00221 & 0.01263 & 0.00220  &  0.00811  &   0.00780  &   0.04480  & 0.00140 & 0.00150 &  0.00257   &   0.00209                     &  0.00892   &  0.00973   \\
\quad LSTM                         & 0.00813 & 0.00652                & 0.01414 & 0.01762 & 0.01406 & 0.01997          & 0.00125 & 0.00930 & 0.00201  & 0.00746   &  0.03392   &  0.02678   & 0.00097 & 0.00143 &  0.00257   &  0.00210                      &  0.00846   &  0.00928   \\
\quad GRU                          & 0.00824 & 0.00654                 & 0.01352 & 0.02116 & 0.01576 & 0.02170           & 0.00145 & 0.00825 &  0.00274 &  0.00729  &  0.03381   &  0.02802   & 0.00110 & 0.00145 &  0.00200   &  0.00202                    &   0.00834  &  0.00908   \\
\quad \textbf{VARNN--RM \ignore{(error memory)}}          & \textbf{0.00412} & \textbf{0.00328}     & \textbf{0.00037} & \textbf{0.00020} & \textbf{0.00041} & \textbf{0.00048}                            & \textbf{0.00016} & \textbf{0.00015} & \textbf{0.00017}  & \textbf{0.00031}   &  \textbf{0.00015}   &  \textbf{0.00024}   & \textbf{0.00024} & \textbf{0.00026} &  \textbf{0.00085}   &   \textbf{0.00058}                                           &  \textbf{0.00028}   &   \textbf{0.00045}   \\
\quad \textbf{VARNN--RM+AM \ignore{(error + activation)}} & \textbf{0.00378} & \textbf{0.00329}       & \textbf{0.00035} & \textbf{0.00018} & \textbf{0.00044} & \textbf{0.00049}                            & \textbf{0.00016} & \textbf{0.00015} & \textbf{0.00017}  &  \textbf{0.00029}  &  \textbf{0.00015}   &  \textbf{0.00022}   & \textbf{0.00022} & \textbf{0.00025}&  \textbf{0.00077}   &  \textbf{0.00057}                                             &  \textbf{0.00037}   &   \textbf{0.00042}  \\

\bottomrule
\end{tabular}%
}
\end{table*}

\textbf{Overall performance.} \textsc{VARNN} consistently achieves the lowest test MSE across all datasets. On \emph{Appliances}, the strongest baseline ARX-LR reaches $5.04\times 10^{-3}$, while \textsc{VARNN-RM} base model improves to $3.28\times 10^{-3}$, marking a $\sim35\%$ reduction. On \emph{HR}, ARX-LR yields $2.7\times 10^{-4}$ test error, but \textsc{VARNN-RM} reduces this to $1.5\times 10^{-4}$, almost halving the error $\sim44.4\%$ . On \emph{PM2.5}, \textsc{VARNN-RM} reaches $2.6\times 10^{-4}$ compared to $4.9\times 10^{-4}$ for ARX-LR, again nearly a twofold improvement $\sim47\%$. In all cases, the gap between training and test MSE remains small, underscoring robustness and resistance to overfitting. VARNN also outperforms the evaluated ARMAX, which is consistent with the benefit of learning a nonlinear residual-conditioning mechanism rather than relying on a fixed linear innovation model.

\textbf{Static vs.\ Dynamic} Static regressors that ignore temporal structure substantially underperform dynamic families on every dataset. RF, in particular, shows pronounced overfitting: near-zero training error but much larger test error. This highlights the need for temporal context.

\textbf{Lags vs.\ Sequence models.} Explicit-lag baselines generally improve over static models but can be sensitive to lag design and model capacity. Classical recurrent baselines deliver mixed results: competitive on \emph{Appliances} and \emph{PM2.5}, but unstable in other tasks\ignore{\emph{BIDMC HR}, where the train–test gap widens sharply}. In contrast, \textsc{VARNN} attains both strong accuracy and stable generalization across domains.

\begin{figure}[t]
  \centering
  \begin{subfigure}[b]{0.45\textwidth}
    \centering
    \includegraphics[width=\textwidth]{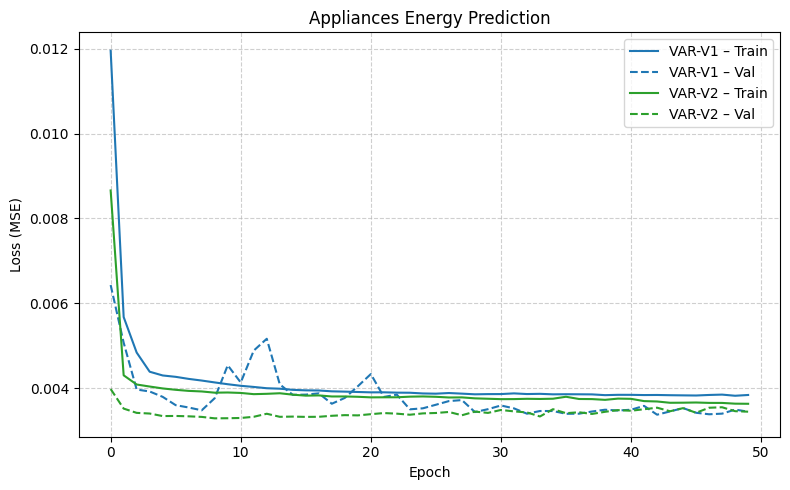}
    \caption{Appliances}
    \label{fig:appliances}
  \end{subfigure}\hfill
  \begin{subfigure}[b]{0.45\textwidth}
    \centering
    \includegraphics[width=\textwidth]{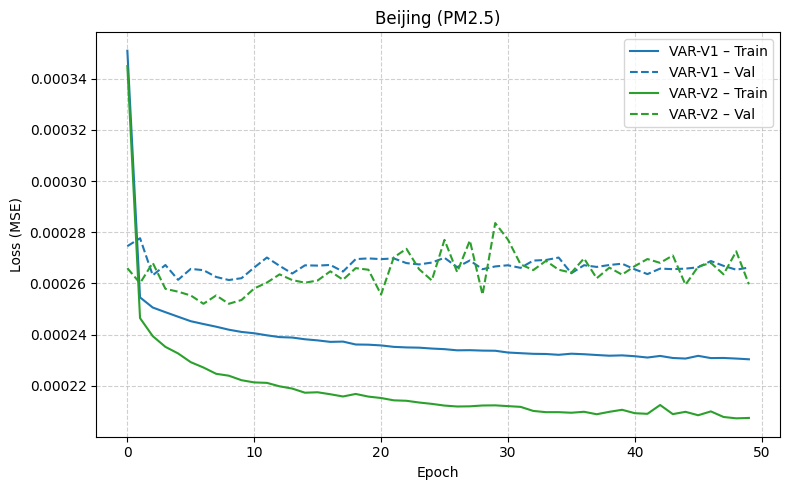}
    \caption{Beijing (PM2.5)}
    \label{fig:pm}
  \end{subfigure}

  \caption{Learning curves for VARNN-RM and VARNN-RM+AM across Appliances and Beijing}
  \label{fig:learning-curves}
\end{figure}

\textbf{VARNN variants: RM vs RM+AM} The learning curve in Figure\ref{fig:learning-curves} shows that the \textsc{RM+AM} variant reveals a faster convergence speed advantage over the base model \textsc{RM} with broadly similar or better generalization. This finding suggests that adding the activation memory (AM), which carries the previous latent activation, enriches short-term temporal dynamics \ignore{representations} and stabilizes predictions.

\subsection{Ablations}
\label{sec:ablation}
We conduct targeted ablations on the residual-memory mechanism in the base model \textbf{VARNN-RM}. All ablations use the same training protocol window length $w$, feature dimension $d$, activation hidden units $k$ and residual memory width $m$, unless otherwise stated. Each configuration is run over the same seed (2025) and reported on MSE loss. Additional analyses of context-window length and and the extension from one-step-ahead to direct multi-step forecasting are provided in Appendices~\ref{app:window_sensitivity}
and~\ref{sec:multistep_forecasting}, respectively.

\par
\begin{figure}[!t]
  \centering

  \begin{subfigure}[b]{0.47\linewidth}
    \centering
    \includegraphics[width=\linewidth]{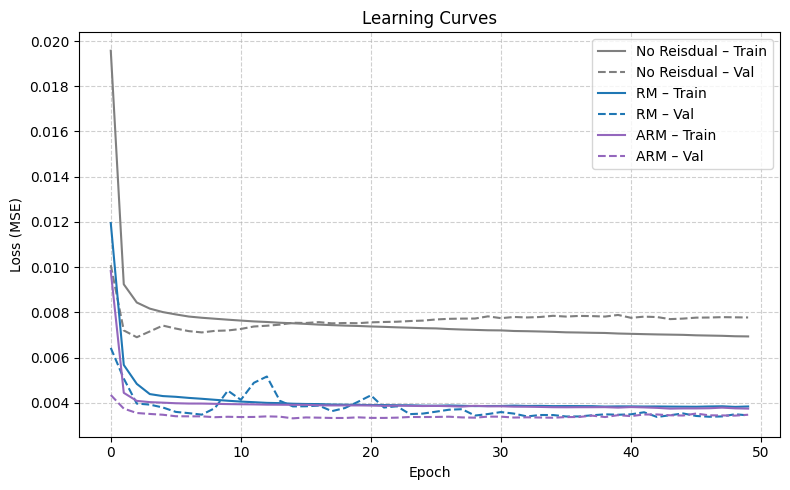}
    \caption{Appliances}
    \label{fig:memoryeffect-appliances}
  \end{subfigure}
  \hfill
  \begin{subfigure}[b]{0.47\linewidth}
    \centering
    \includegraphics[width=\linewidth]{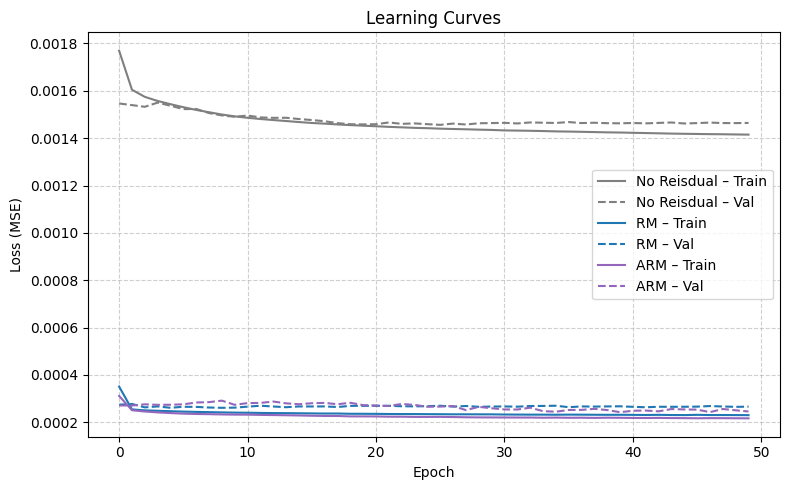}
    \caption{Beijing (PM2.5)}
    \label{fig:memoryeffect-pm}
  \end{subfigure}

  \caption{Learning curves for VARNN-RM and RM+AM across Appliances and Beijing datasets.}
  \label{fig:memory-effect}
\end{figure}

\medskip
\textbf{Effect of Residual Feedback}
Figure~\ref{fig:memory-effect} compares training and validation learning curves for three variants: 
(i) \emph{no residual} (the predictor receives a zero residual state), 
(ii) \emph{residual memory (RM)}, and 
(iii) \emph{accumulative residual memory (ARM)}. Across both PM2.5 and Appliances, residual-aware models converge more quickly and attain substantially lower test MSE than the baseline. RM alone already delivers most of the improvement, while ARM yields marginal additional gains and stability. The results confirm that feeding back prediction errors as a learnable memory state improves stability and generalization compared to models that absorb variability only implicitly.

\begin{figure*}[!t]
  \centering
  \begin{subfigure}[b]{0.33\textwidth}
    \centering
    \includegraphics[width=\textwidth]{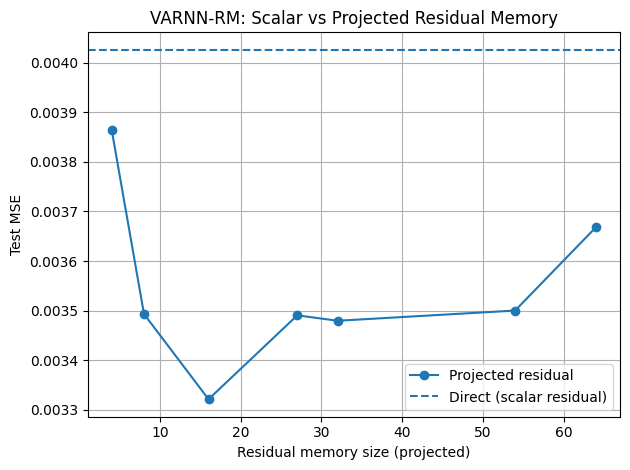}
    \caption{Appliances}
    \label{fig:appliances}
  \end{subfigure}\hfill
  \begin{subfigure}[b]{0.33\textwidth}
    \centering
    \includegraphics[width=\textwidth]{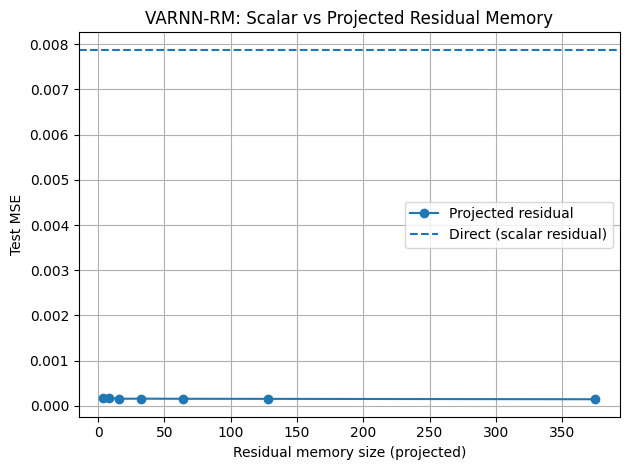}
    \caption{BIDMC (HR)}
    \label{fig:hr}
  \end{subfigure}\hfill
  \begin{subfigure}[b]{0.33\textwidth}
    \centering
    \includegraphics[width=\textwidth]{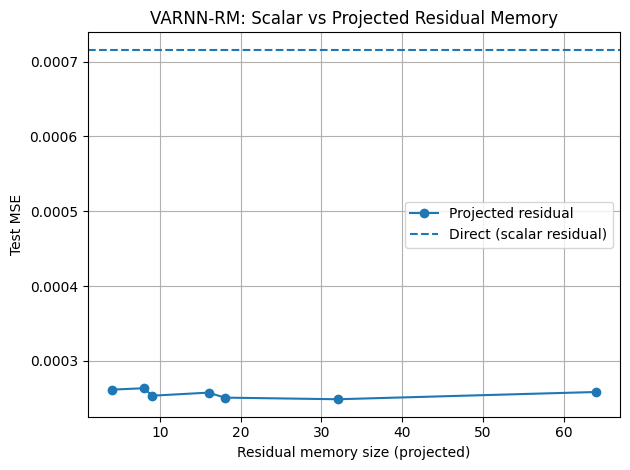}
    \caption{Beijing (PM2.5)}
    \label{fig:pm}
  \end{subfigure}

  \caption{Residual memory size vs scalar across three datasets.}
  \label{fig:memory-size}
\end{figure*}

\textbf{Residual-memory: scalar vs.\ projected} \label{par:rm-scalar-vs-proj}
We compare raw scalar feedback $h_\tau=e_\tau$ with the projected state $\mathbf{h}_\tau=\rho(\mathbf{W}_\epsilon e_\tau+\mathbf{b}_\epsilon)$ in VARNN-RM. The predictor width and other training settings are fixed while $m\in\{4,8,16,32,64,\min(128,2d),d\}$ varies. As shown in \autoref{fig:memory-size}, projecting the residual consistently outperforms the scalar variant:
\emph{Appliances:} scalar $0.004026$ vs.\ best projected $0.003321$ ($m{=}16$), \textbf{17.5\%} lower; \emph{BIDMC HR:} scalar $0.007978$ vs.\ best $0.000154$ ($m{=}64$), \textbf{98.1\%} lower;
\emph{Beijing PM2.5:} scalar $0.000702$ vs.\ best $0.000246$ ($m{=}4$), \textbf{65.0\%} lower.
These results indicate that a scalar feedback cannot represent diverse error regimes and may cause noise, thereby downgrading the model's performance. In contrast, a learned vector memory provides a richer residual representation and yields lower error. The optimal $m$ is dataset dependent: small on PM2.5, moderate on Appliances, and larger on BIDMC due to richer waveform dynamics.

\section{Ethics statement}
This work develops supervised learning methods for time-series regression. Ethical risks are low and domain-dependent. When applied to sensitive domains (e.g., healthcare), appropriate consent, privacy safeguards, and fairness audits should be considered.

\section{Conclusion and Future Work}

We introduced the Variability-Aware Recursive Neural Network
(\textsc{VARNN}), a residual-aware model for dynamic time-series
regression that maps scalar prediction innovations into a learned
nonlinear residual-memory state and reuses this representation
for subsequent prediction. Across nine datasets spanning energy,
healthcare, and environmental domains, VARNN achieves lower test
MSE than the compared static, lag-based, and sequence-model
baselines. Targeted ablations further show that non-linear latent residual memory improves on direct scalar residual feedback, supporting the importance of explicitly learning how prediction deviations are represented and reused. While this work focuses primarily on short-context supervised
prediction with observed-outcome feedback, the residual-memory
principle may also be useful in other temporal learning tasks.
Future work will investigate VARNN-based representations for
anomaly detection, where evolving residual states may help
characterize departures from normal dynamics, and for irregularly
sampled time series, where memory updates could explicitly account
for variable intervals between observations.

\newpage
\bibliography{references}
\bibliographystyle{iclr2027_conference}

\newpage
\appendix

\section{Disclosure of AI Assistance}
All conceptual development, modeling, experiments, and analysis were carried out by the authors. Portions of the writing were refined using large language models (LLMs), which supported grammar, clarity, and formatting.

\section{Theoretical Justification}
\label{sec:theory}

Given supervised time–series regression data with multivariate covariates and a univariate response:
$\{(\mathbf{x}_\tau$, and $ y_\tau)\}_{\tau=1}^{T}$, where $ \mathbf{x}_\tau \in \mathbb{R}^d,\; y_\tau \in \mathbb{R}.$
We consider one-step-ahead prediction with covariates under temporal
variability. Under squared loss, let
\begin{equation}
f^\star(\mathbf{x}_t)
:=
\mathbb{E}[y_t \mid \mathbf{x}_t]
\label{eq:optimal_covariate}
\end{equation}
denote the ideal prediction rule based only on the currently observed
covariates. The target can then be written as
\begin{equation}
y_t
=
f^\star(\mathbf{x}_t)
+
e_t
\label{eq:decomp_theory}
\end{equation}
where \(f^\star(\mathbf{x}_t)\) denotes the ideal covariate-driven
component and \(e_t\) is the residual term. This residual term may
reflect drift, non-stationarity, regime shifts, heteroscedasticity, or
other unobserved temporal effects that cannot be explained by
\(\mathbf{x}_t\) alone.

By definition,
\begin{equation}
e_t
=
y_t-f^\star(\mathbf{x}_t)
\end{equation}
and therefore
\begin{equation}
\mathbb{E}[e_t \mid \mathbf{x}_t]
=
0
\label{eq:zero_residual}
\end{equation}

Thus, when only the current covariates are available, the optimal
one-step predictor is
\begin{equation}
\mathbb{E}[y_t \mid \mathbf{x}_t]
=
f^\star(\mathbf{x}_t)
\end{equation}

However, under temporal variability, information contained in recent
prediction errors may provide additional information about the
next-step residual. Let \(\mathbf{h}_{t-1}\) denote the residual-memory
state constructed from previously observed prediction errors. Although
\(e_t\) is unpredictable from \(\mathbf{x}_t\) alone in expectation, it
may remain predictable when the residual-memory state is also observed,
such that
\begin{equation}
\mathbb{E}
[e_t \mid \mathbf{x}_t,\mathbf{h}_{t-1}]
\neq 0
\label{eq:residual_predictable}
\end{equation}

Using \eqref{eq:decomp_theory}, the optimal predictor conditioned on
both the current covariates and the residual-memory state becomes
\begin{align}
\mathbb{E}
[y_t \mid \mathbf{x}_t,\mathbf{h}_{t-1}]
&=
\mathbb{E}
\left[
f^\star(\mathbf{x}_t)+e_t
\mid
\mathbf{x}_t,\mathbf{h}_{t-1}
\right] \\
&=
f^\star(\mathbf{x}_t)
+
\mathbb{E}
[e_t \mid \mathbf{x}_t,\mathbf{h}_{t-1}]
\label{eq:conditional_correction}
\end{align}

\Eqref{eq:conditional_correction} shows that the prediction can
be interpreted as the sum of a covariate-driven component and a
residual correction induced by information contained in the
residual-memory state. When the conditional residual term is nonzero,
\(\mathbf{h}_{t-1}\) provides predictive information beyond the current
covariates alone.
Under squared loss, conditioning on this additional information cannot
increase the minimum achievable expected prediction error. Since
\begin{equation}
\sigma(\mathbf{x}_t)
\subseteq
\sigma(\mathbf{x}_t,\mathbf{h}_{t-1})
\end{equation}

we have
\begin{equation}
\mathbb{E}
\left[
\left(
y_t-
\mathbb{E}[y_t\mid
\mathbf{x}_t,\mathbf{h}_{t-1}]
\right)^2
\right]
\leq
\mathbb{E}
\left[
\left(
y_t-
\mathbb{E}[y_t\mid\mathbf{x}_t]
\right)^2
\right]
\label{eq:risk_reduction}
\end{equation}

Moreover, the inequality is strict whenever the residual-memory state
contains additional predictive information, i.e.,
\begin{equation}
\mathbb{P}
\left(
\mathbb{E}
[y_t\mid\mathbf{x}_t,\mathbf{h}_{t-1}]
\neq
\mathbb{E}
[y_t\mid\mathbf{x}_t]
\right)
>0
\label{eq:strict_improvement}
\end{equation}

Therefore, residual memory is theoretically motivated when recent
prediction errors contain structured information about the next-step
residual that is not already captured by the current covariates.
VARNN is designed for this setting by learning a residual-memory
representation from previous prediction errors and incorporating this
state into the subsequent prediction. In this way, the residual-memory
pathway provides a mechanism for exploiting predictable temporal
deviations that would otherwise remain unmodeled by a covariate-only
predictor.

\subsection{Sensitivity to Window Length}
\label{app:window_sensitivity}

In the main experiments, we use a context window of length $w=5$.
To examine whether VARNN is sensitive to this choice, we evaluate the
base VARNN-RM model using $w \in \{2,3,5,7,9\}$ on three representative
datasets from different domains: Appliances (energy), BIDMC-HR
(healthcare), and Beijing PM2.5 (environmental monitoring). All other
model and training settings are kept unchanged.

Table~\ref{tab:window_sensitivity} reports the test MSE for each window
length. Overall, VARNN remains competitive across the evaluated
short-context lengths, indicating that its performance is reasonably
stable with respect to moderate changes in $w$. Importantly, $w=5$
achieves the lowest test MSE across all three representative datasets,
while both shorter and longer windows remain broadly competitive.
Increasing the context beyond $w=5$ does not provide consistent
improvements and can slightly degrade performance, particularly on
BIDMC-HR.

These results support the use of $w=5$ as a common short-context setting
in the main experiments rather than a dataset-specific choice. They are
also consistent with the intended role of VARNN as a short-context
residual-memory model, where a small number of recent prediction deviations
appears sufficient to provide an informative correction signal, while
increasing the historical context does not yield consistent improvements.
This suggests that VARNN can effectively exploit local residual dynamics
without requiring long-range sequence memory.

\begin{table}[t]
\centering
\caption{Sensitivity of VARNN-RM to the context window length $w$.
Test MSE is reported; lower is better. Best results for each dataset
are shown in bold.}
\label{tab:window_sensitivity}
\small
\setlength{\tabcolsep}{6pt}
\begin{tabular}{c c c c}
\toprule
\textbf{$w$} &
\textbf{Appliances} &
\textbf{BIDMC-HR} &
\textbf{Beijing PM2.5} \\
\midrule
2 & 0.00338 & 0.00016 & 0.00028 \\
3 & 0.00358 & 0.00018 & 0.00029 \\
5 & \textbf{0.00328} & \textbf{0.00015} & \textbf{0.00026} \\
7 & 0.00340 & 0.00016 & \textbf{0.00026} \\
9 & 0.00340 & 0.00020 & 0.00027 \\
\bottomrule
\end{tabular}
\end{table}

\section{Representation Learning of Variability in VARNN}

\textsc{VARNN} learns a representation of variability by encoding each one step prediction innovation $e_{\tau} = y_{\tau} - \hat{y}_{\tau}$ into an multi-dimensional residual memory vector \(\mathbf{h}_{\tau}\in\mathbb{R}^{m}\) (Eq.~\ref{eq:memory}, \ref{eq:memory-ARM}). Rather than passing the innovation as a raw scalar as used approach in the ARMAX model or error injection heuristics models, \textsc{VARNN} projects the residual through a learnable transformation that captures richer structure:
\begin{flalign}
& \mathbf{h}_{\tau} = \rho\!\left(\mathbf{W}_\epsilon e_{\tau} + \mathbf{b}_\epsilon\right)
&& \text{(Residual Memory (RM) variant)} \nonumber\
\end{flalign}
In the accumulative variant (ARM), this embedding composes recursively with the previous memory state, forming a learned autoregressive filter on recent deviations:
\begin{flalign}
& \mathbf{h}_\tau = \rho\!\left(W_h \mathbf{h}_{\tau-1} + W_\epsilon e_\tau + b_\epsilon\right)
&& \text{(Accumulative Residual Memory (ARM) variant)} \nonumber\
\end{flalign}

This projection allows the residual memory to encode richer properties of variability that a scalar cannot express such as magnitude, sign asymmetry, heteroscedastic volatility, burstiness, and short-term bias or drift regimes. The resulting memory state is then fed back into the predictive pathway via the fusion vector \(\mathbf{z}_{\tau}\) (Eq.~\ref{eq:fusion}), enabling each next prediction to be conditioned on a learned summary of recent deviations.

\par\medskip
In contrast, models that directly ingest the raw error as an additional input feature have no mechanism to shape or filter the noise structure of innovations, which empirically leads to noise amplification and performance degradation. For example, under identical training protocol and windowing size w=5 , ARMAX--LR exhibits substantially worse accuracy across datasets, most notably on ETTh1  (\textbf{0.02952} vs.\ \textsc{VARNN}–RM’s \textbf{0.00020}) and  ETTh2 (\textbf{0.14067} vs.\ \textsc{VARNN}–RM’s \textbf{.00048}), showing that scalar innovation injection amplifies noise rather than providing a stable conditioning  signal (see Results, \autoref{tab:main-results}). Similar degradation appears in healthcare and  environmental datasets, where ARMAX--LR consistently ranks among the weakest performers.

\par\medskip
As shown in our ablation study (\S\ref{par:rm-scalar-vs-proj}), using a scalar residual state yields substantially worse test MSE across all three domains for example, replacing the scalar with a projected memory vector reduces error by \(17.5\%\) on Appliances, \(65\%\) on PM2.5 and \(98\%\) on BIDMC HR. Thus, \textsc{VARNN} does not simply reuse prediction errors, it learns a structured residual representation that filters noise, stabilizes optimization, and preserves meaningful variability patterns for next-step prediction. This makes \textsc{VARNN} fundamentally distinguishing from ARMAX, ARX-style or recurrent models that absorb variability only implicitly, providing a principled and empirically validated approach to handling non-stationarity.

\section{Additional Implementation and Analysis}
The main paper establishes that residual memory improves robustness under non-stationarity. This appendix provides complementary implementation details and extended analyses to support the main results presented in Section~\ref{sec:results}. We first investigate the effect of the activation function within the residual-memory block (\S\ref{app:relu-vs-tanh}), comparing \textsc{ReLU} and \textsc{tanh} activations to understand their impact on convergence stability and noise sensitivity. We then analyze the behavior of the four VARNN variants: \textsc{RM}, \textsc{RM+AM}, \textsc{ARM}, and \textsc{ARM+AM} on the Beijing PM2.5 dataset (\S\ref{app:pm25-variants}), illustrating how residual accumulation and activation carry jointly affect temporal adaptation under non-stationary dynamics. Finally, we provide algorithmic and quantitative summaries of all configurations (\S\ref{app:alg-view}, \S\ref{app:varnn-variants-results}), including detailed pseudocode and per-variant performance metrics across datasets. Together, these results offer a deeper understanding of VARNN’s internal mechanisms, training behavior, and architectural trade-offs, reinforcing its design rationale and empirical consistency across diverse temporal regimes.

\section{Activation Function in Residual-Memory Block (ReLU vs.\ Tanh)}
\label{app:relu-vs-tanh}

\begin{figure}[h]
  \centering
  \begin{subfigure}[b]{0.5\textwidth}
    \centering
    \includegraphics[width=\textwidth]{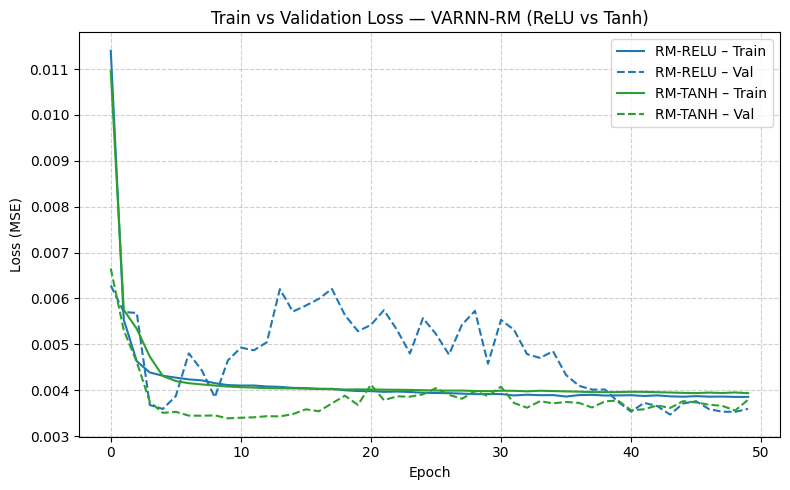}
    \caption{Appliances}
    \label{fig:appliances}
  \end{subfigure}\hfill
  \begin{subfigure}[b]{0.5\textwidth}
    \centering
    \includegraphics[width=\textwidth]{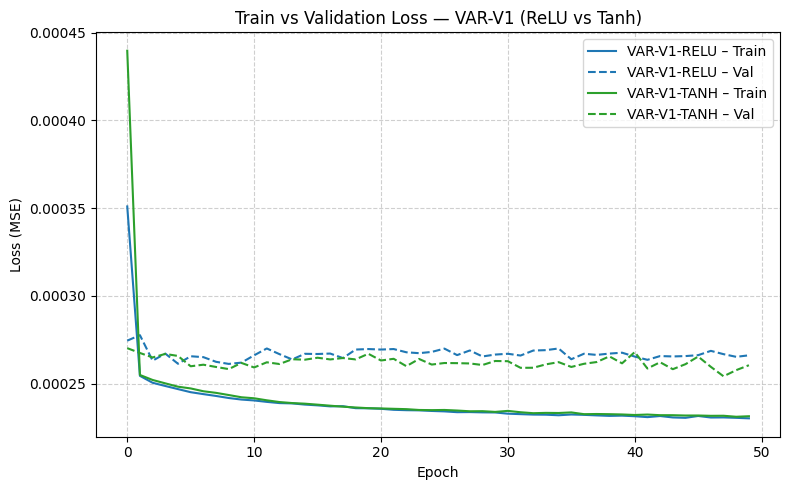}
    \caption{Beijing (PM2.5)}
    \label{fig:pm}
  \end{subfigure}
  \caption{Train vs.\ validation loss on \emph{Beijing PM2.5} using VARNN-RM with ReLU vs.\ Tanh activations.}
  \label{fig:activation-function}
\end{figure}

We conducted an ablation to examine the effect of the residual-memory activation $\rho(\cdot)$ in VARNN-RM, comparing the default \textsc{ReLU} against a bounded \textsc{tanh}. Results are reported on two representative datasets: \emph{Appliances Energy Prediction} and \emph{Beijing PM2.5}. The learning curves graphs in Figure~\ref{fig:activation-function} show the train/validation learning curves over 50 epochs.

On \textbf{Appliances}, both activations converge to a similar validation floor ($\approx\!3.9{\times}10^{-3}$ MSE), but \textsc{tanh} exhibits smoother trajectories with reduced oscillations near convergence. In contrast, \textsc{ReLU} shows higher variance across epochs, suggesting sensitivity to local error bursts.  

On \textbf{PM2.5}, the difference is negligible: both activations stabilize around $2.6{\times}10^{-4}$ MSE, with \textsc{tanh} again displaying slightly lower validation variance. The bounded range of \textsc{tanh} likely dampens extreme residual updates, which can be beneficial under noisy or volatile conditions.  

Overall, these results indicate that while \textsc{ReLU} and \textsc{tanh} achieve comparable final accuracy, \textsc{tanh} provides smoother convergence and more stable generalization in non-stationary settings. Unless sparsity in the residual embedding is explicitly desired, we recommend \textsc{tanh} as a default activation in the residual-memory block.


\subsection{Variants Performance on Beijing PM2.5}
\label{app:pm25-variants}
\begin{figure}[t]
  \centering
  \includegraphics[width=0.85\linewidth]{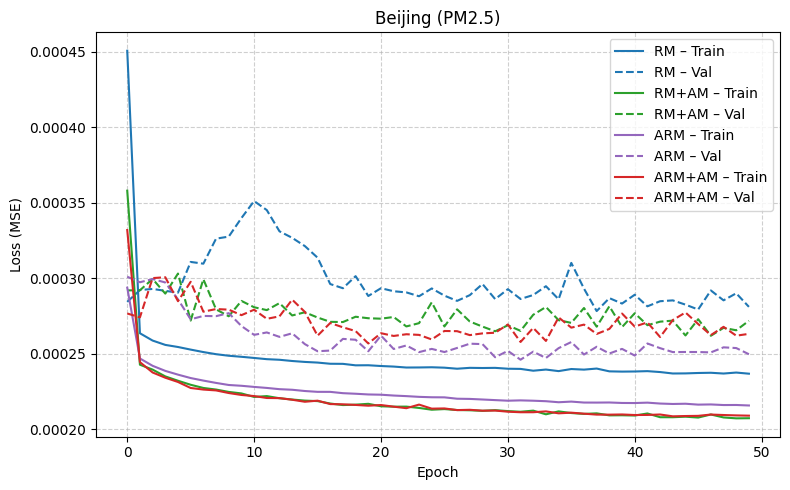}
  \caption{Beijing PM2.5: learning curves for VARNN variants. Accumulating residuals (ARM) improves stability and validation accuracy, while combining accumulation with activation memory (ARM+AM)  yields similar performance with slightly smoother training dynamics.}
  \label{fig:varnn-pm}
\end{figure}

Figure~\ref{fig:varnn-pm} compares training and validation learning curves across the four VARNN variants: \emph{residual memory (RM)}, \emph{residual+activation memory (RM+AM)}, \emph{accumulative residual memory (ARM)}, and \emph{ARM+AM}: 

\textbf{Residual memory (RM).} The base model VARNN-RM converges steadily but exhibits the highest validation loss among variants, with noticeable fluctuations during the first 20 epochs. This indicates that a pure residual pathway provides stability but underfits the variability in PM2.5 dynamics.

\textbf{RM+AM.} Adding activation memory improves training convergence but validation loss remains noisy, suggesting that activation carry alone does not robustly capture nonstationary fluctuations in air-quality data.

\textbf{ARM.} ARM achieves Lowest validation loss with smooth convergence after $\sim$10--15 epochs, demonstrating that the accumulating residuals are crucial on PM2.5. This demonstrates the benefit of accumulating residuals over time to encode volatility regimes.

\textbf{ARM+AM.} Training loss is the lowest, and validation performance closely tracks ARM, typically marginally higher or delayed by a few epochs.

In summary, on Beijing PM2.5, \textbf{ARM} achieves the best validation stability, with \textbf{ARM+AM} performing comparably. Therefore, the key driver is residual \emph{accumulation}; activation offers little additional benefit for this dataset.

\clearpage
\subsection{Multi-Step Forecasting}
\label{sec:multistep_forecasting}

The original VARNN formulation focuses on one-step-ahead forecasting with covariates, and demonstrates improved one-step-ahead forecasting performance over conventional recurrent models as shown in Table~\ref{tab:main-results} of the main paper. To evaluate whether the benefit of VARNN's residual-aware encoder persists beyond 
the immediate next-step prediction, we extend VARNN to direct multi-step forecasting with horizons
$H \in \{1,3,5,7\}$ and directly predict all future steps from the learned representation using a linear prediction head, while using the same historical window length ($w{=}5$) and forecast horizons across models.
This experiment evaluates whether the representation learned by VARNN remains effective as the forecasting horizon increases.

Table~\ref{tab:multistep_forecasting} shows that the VARNN variants consistently outperform the conventional RNN, GRU, and LSTM
baselines across all datasets and forecasting horizons. 
Notably, the base VARNN-RM model already achieves lower MSE than the recurrent baselines across all evaluated horizons, while the additional VARNN variants further improve performance depending on the dataset and horizon.
Relative to the strongest conventional recurrent baseline at each horizon, the best VARNN variant
reduces MSE by approximately $96.8\%$--$98.6\%$ on HR, $35.5\%$--$83.3\%$ on PM2.5, and $18.6\%$--$48.2\%$ on Power.
Overall, these results suggest that the residual-aware representation learned by the VARNN encoder remains effective
beyond the original one-step-ahead forecasting setting and provides useful predictive information for multi-step forecasting.

\begin{table}[h]
\centering
\caption{
Direct multi-step forecasting performance in MSE for prediction horizons
$H \in \{1,3,5,7\}$. Lower is better. All models use the same historical window length ($w{=}5$)  and direct multi-step prediction protocol. The best result for each
dataset and horizon is shown in bold.
}
\label{tab:multistep_forecasting}
\resizebox{\textwidth}{!}{
\begin{tabular}{lcccccccccccc}
\toprule
& \multicolumn{4}{c}{Power}
& \multicolumn{4}{c}{HR}
& \multicolumn{4}{c}{PM2.5} \\
\cmidrule(lr){2-5}
\cmidrule(lr){6-9}
\cmidrule(lr){10-13}
Model
& $H{=}1$ & $H{=}3$ & $H{=}5$ & $H{=}7$
& $H{=}1$ & $H{=}3$ & $H{=}5$ & $H{=}7$
& $H{=}1$ & $H{=}3$ & $H{=}5$ & $H{=}7$ \\
\midrule

RNN
& 0.007057 & 0.007459 & 0.007310 & 0.007337
& 0.010834 & 0.019918 & 0.017077 & 0.016677
& 0.001407 & 0.001773 & 0.001995 & 0.002268 \\

GRU
& 0.006663 & 0.006711 & 0.006870 & 0.006772
& 0.016132 & 0.017971 & 0.017825 & 0.015146
& 0.001412 & 0.001765 & 0.002015 & 0.002303 \\

LSTM-H
& 0.006522 & 0.006510 & 0.006793 & 0.006731
& 0.017026 & 0.018671 & 0.016860 & 0.017115
& 0.001365 & 0.001747 & 0.001984 & 0.002257 \\

LSTM-H+C
& 0.006484 & 0.007736 & 0.007053 & 0.006923
& 0.017054 & 0.017482 & 0.018643 & 0.016395
& 0.001408 & 0.001757 & 0.002021 & 0.002260 \\

\midrule

VARNN-RM
& 0.003430 & 0.005029 & 0.005817 & 0.006147
& 0.000153 & 0.000335 & 0.000397 & \textbf{0.000483}
& 0.000230 & 0.000656 & 0.001082 & 0.001493 \\

VARNN-ARM
& 0.003399 & 0.005082 & 0.005789 & 0.005932
& 0.000164 & 0.000390 & 0.000387 & 0.000567
& \textbf{0.000228} & \textbf{0.000628} & 0.001064 & \textbf{0.001456} \\

VARNN-AMRM
& 0.003499 & 0.004916 & \textbf{0.005292} & \textbf{0.005477}
& \textbf{0.000149} & 0.000366 & 0.000387 & 0.000490
& 0.000236 & 0.000631 & \textbf{0.001050} & 0.001466 \\

VARNN-AMARM
& \textbf{0.003358} & \textbf{0.004835} & 0.005373 & 0.005537
& 0.000163 & \textbf{0.000321} & \textbf{0.000383} & 0.000566
& 0.000229 & 0.000636 & \textbf{0.001050} & 0.001467 \\

\bottomrule
\end{tabular}
}
\end{table}

\section{Extended Results and Figures}
\subsection{Algorithmic View Over All VARNN Variants}
\label{app:alg-view}
Algorithm~\ref{alg:varnn} summarizes the teacher-forced unroll over the $(w{-}1)$ labeled context steps and the current-time prediction at $t$. We use $\theta(\cdot)$ for the predictor nonlinearity (e.g., ReLU) and $\rho(\cdot)$ for the residual-memory update (ReLU or $\tanh$). The four configurations correspond to: 
\textbf{RM} (residual memory only), 
\textbf{RM+AM} (residual + activation memory), 
\textbf{ARM} (accumulative residual memory), and 
\textbf{ARM+AM} (accumulative residual + activation memory).

\noindent\textbf{Notes.}
(i) RM and RM+AM use an \emph{instantaneous residual memory} update $\mathbf{h}_\tau \!\leftarrow\! \rho(\mathbf{W}_r e_\tau+\mathbf{b}_r)$.  
(ii) ARM and ARM+AM use an \emph{accumulative residual memory} update $\mathbf{h}_\tau \!\leftarrow\! \rho(\mathbf{W}_r e_\tau+\mathbf{W}_h \mathbf{h}_{\tau-1}+\mathbf{b}_r)$.  
(iii) RM+AM and ARM+AM additionally carry the previous activation $\mathbf{u}_{\tau-1}$ forward as part of the state.  
(iv) $\theta(\cdot)$ is typically ReLU; $\rho(\cdot)$ can be ReLU or $\tanh$ depending on the ablation.

\subsection{Performance of VARNN Variants}
\label{app:varnn-variants-results}

Table~\ref{tab:varnn-variants} reports the train and test MSE for all four VARNN configurations introduced in Section~\ref{sec:variants}: 
\textbf{RM}, \textbf{RM+AM}, \textbf{ARM}, and \textbf{ARM+AM}. 
While the accumulative variants (\textbf{ARM}, \textbf{ARM+AM}) achieve slightly lower errors on datasets with slow drift (e.g., PM2.5), the instantaneous residual-memory models (\textbf{RM}, \textbf{RM+AM}) perform comparably across all domains with lower computational cost. 
For clarity, only the configuration (VARNN–RM or VARNN–RM+AM) is reported in the main results table (Table~\ref{tab:main-results}).

\begin{table}[h]
\centering
\caption{Train and test mean squared error (MSE; ↓) on all datasets. 
\emph{Static} models use only contemporaneous covariates $\mathbf{x}_t$; 
\emph{Dynamic} models additionally incorporate lagged outputs/inputs or recurrent states. 
Best test scores are in \textbf{bold}.}
\label{tab:varnn-variants}
\setlength{\tabcolsep}{3pt}

\resizebox{\textwidth}{!}{%
\renewcommand{\arraystretch}{1.4}

\begin{tabular}{lcc|cc|cc|cc|cc|cc|cc|cc|cc}
\toprule

\multirow{3}{*}{\textbf{Model}} 
& \multicolumn{6}{c}{\textbf{ENERGY}}
& \multicolumn{6}{c}{\textbf{Healthcare}}
& \multicolumn{6}{c}{\textbf{Environmental}} \\
\cmidrule(lr){2-7}\cmidrule(lr){8-13}\cmidrule(lr){14-19}
& \multicolumn{2}{c}{\textbf{Appliances}} 
& \multicolumn{2}{c}{\textbf{ETTh1}}
& \multicolumn{2}{c}{\textbf{ETTh2}} 
& \multicolumn{2}{c}{\textbf{BIDMC HR}} 
& \multicolumn{2}{c}{\textbf{BIDMC RR}} 
& \multicolumn{2}{c}{\textbf{BIDMC SPO2}} 
& \multicolumn{2}{c}{\textbf{Beijing PM2.5}}
& \multicolumn{2}{c}{\textbf{Beijing PM10}}

& \multicolumn{2}{c}{\textbf{WEATHER}} 
\\
\cmidrule(r){2-3} \cmidrule(r){4-5} \cmidrule(r){6-7} \cmidrule(r){8-9} \cmidrule(r){10-11} \cmidrule(r){12-13}  \cmidrule(r){14-15}  \cmidrule(r){16-17} \cmidrule(r){18-19}
& \textbf{Train} & \textbf{Test} & \textbf{Train} & \textbf{Test} & \textbf{Train} & \textbf{Test} & \textbf{Train} & \textbf{Test} & \textbf{Train} & \textbf{Test} & \textbf{Train} & \textbf{Test} & \textbf{Train}   & \textbf{Test} & \textbf{Train}&\textbf{Test} & \textbf{Train} &\textbf{Test}\\
\midrule

VARNN--RM          & 0.00412 & 0.00328    & 0.00037 & 0.00020 & 0.00041 & 0.00048 & 0.00016 & 0.00015 & 0.00017  & 0.00031  &  0.00015   &  0.00024  & 0.00024 & 0.00026 &  0.00085   &  0.00058  &  0.00028   &   0.00045   \\
VARNN--RM+AM       & 0.00378 & 0.00329 & 0.00035 & 0.00018 & 0.00044 & 0.00049 & 0.00016 & 0.00015 & 0.00017 &  0.00029  & 0.00015   &  0.00022 & 0.00022 & 0.00025&  0.00077   &  0.00057  &  0.00037 &  0.00042 \\
VARNN--ARM        & 0.00382 & 0.00330 & 0.00035 & 0.00017 & 0.00040 & 0.00046 & 0.00015 & 0.00015 &  0.00017  &  0.00029  &  0.00016   &  0.00024   & 0.00022 & 0.00025 &  0.00077  &  0.00058 &  0.00026   &  0.00031   \\
VARNN--RM+ARM      & 0.00374 & 0.00329 & 0.00037 & 0.00017 & 0.00041 & 0.00051 & 0.00016 & 0.00015 & 0.00017  &  0.00029  &  0.00015   &  0.00024   & 0.00021 & 0.00024 &  0.00078   &   0.00055  &  0.00024   &   0.00032  \\

\bottomrule
\end{tabular}%
}
\end{table}

As shown, the differences across variants are modest (typically within 1–2\% MSE), confirming that the core residual-memory mechanism is the dominant contributor to VARNN’s robustness. Therefore, the base VARNN–RM is used as the representative model in the main comparisons.

\FloatBarrier
\begin{algorithm}[H]
\caption{VARNN (RM, RM+AM, ARM, ARM+AM): Windowed prediction at time $t$}
\label{alg:varnn}
\small
\begin{algorithmic}[1]
\REQUIRE Window $\mathcal{W}_t=\{(\mathbf{x}_{t-w+1},y_{t-w+1}),\ldots,(\mathbf{x}_{t-1},y_{t-1}),\mathbf{x}_t\}$; configuration $\in\{\text{RM},\text{RM+AM},\text{ARM},\text{ARM+AM}\}$; parameters $\theta=\{\mathbf{W}_z,\mathbf{b}_z,\mathbf{W}_o,\mathbf{b}_o,\mathbf{W}_r,\mathbf{W}_h,\mathbf{b}_r\}$
\ENSURE Current prediction $\hat{y}_t$
\STATE Initialize residual memory $\mathbf{h}_{t-w}\gets \mathbf{0}$
\IF{configuration $\in \{\text{RM+AM},\text{ARM+AM}\}$}
    \STATE Initialize activation memory $\mathbf{u}_{t-w}\gets \mathbf{0}$
\ENDIF
\FOR[teacher-forced context steps]{$\tau=t-w+1$ \textbf{to} $t-1$}
    \IF{configuration $\in \{\text{RM},\text{ARM}\}$}
        \STATE $\mathbf{z}_\tau \gets [\,\mathbf{x}_\tau;\,\mathbf{h}_{\tau-1}\,]$
    \ELSE
        \STATE $\mathbf{z}_\tau \gets [\,\mathbf{x}_\tau;\,\mathbf{h}_{\tau-1};\,\mathbf{u}_{\tau-1}\,]$
    \ENDIF
    \STATE $\mathbf{u}_\tau \gets \theta(\mathbf{W}_z\mathbf{z}_\tau+\mathbf{b}_z)$
    \STATE $\hat{y}_\tau \gets \mathbf{W}_o\mathbf{u}_\tau+\mathbf{b}_o$
    \STATE $e_\tau \gets y_\tau-\hat{y}_\tau$
    \IF{configuration $\in \{\text{RM},\text{RM+AM}\}$}
        \STATE $\mathbf{h}_\tau \gets \rho(\mathbf{W}_r e_\tau+\mathbf{b}_r)$
    \ELSE[ARM / ARM+AM]
        \STATE $\mathbf{h}_\tau \gets \rho(\mathbf{W}_r e_\tau+\mathbf{W}_h\mathbf{h}_{\tau-1}+\mathbf{b}_r)$
    \ENDIF
\ENDFOR
\IF{configuration $\in \{\text{RM},\text{ARM}\}$}
    \STATE $\mathbf{z}_t \gets [\,\mathbf{x}_t;\,\mathbf{h}_{t-1}\,]$
\ELSE
    \STATE $\mathbf{z}_t \gets [\,\mathbf{x}_t;\,\mathbf{h}_{t-1};\,\mathbf{u}_{t-1}\,]$
\ENDIF
\STATE $\hat{y}_t \gets \mathbf{W}_o\,\sigma(\mathbf{W}_z\mathbf{z}_t+\mathbf{b}_z)+\mathbf{b}_o$
\STATE \textbf{return} $\hat{y}_t$ 
\end{algorithmic}
\end{algorithm}


\end{document}